\documentclass[letterpaper, 10 pt, conference]{ieeeconf}
\IEEEoverridecommandlockouts                                       
\usepackage{siunitx}
\usepackage{config/noel}

\title{\LARGE \bf
Planar Bipedal Locomotion with Nonlinear Model Predictive Control: \\ Online Gait Generation using Whole-Body Dynamics
}
\author{Manuel Y. Galliker$^{1\dagger}$, Noel Csomay-Shanklin$^{2\dagger}$, Ruben Grandia$^1$, Andrew J. Taylor$^2$, \\ Farbod Farshidian$^1$, Marco Hutter$^1$, Aaron D. Ames$^2$
\thanks{$^\dagger$ These authors contributed equally to this work.
}
\thanks{$^1$ M. Y. Galliker, R. Grandia, F. Farshidian, and M. Hutter are with the Department of Mechanical and Process Engineering, ETH Z\"{u}rich, 8092 Z\"{u}rich, Switzerland
        {\tt\footnotesize \{manuelga, rgrandia, farbodf, mahutter\}@ethz.ch}
        {\tt\small}}%
\thanks{$^2$ N. Csomay-Shanklin, A. J. Taylor, and A. D. Ames are with the Department of Computing and Mathematical Sciences, California Institute of Technology, Pasadena, CA, 91125, USA
        {\tt\footnotesize \{noelcs, ajtaylor, ames\}@caltech.edu}
        {\tt\small}}%
\thanks{This research was supported by NSF NRI award 1924526, NSF award 1932091, NSF CMMI award 1923239, and the Swiss National Science Foundation through the National Centre of Competence in Research Robotics (NCCR Robotics).}%
}

\begin{document}
\maketitle
\thispagestyle{empty}
\pagestyle{empty}

\begin{abstract}
The ability to generate dynamic walking in real-time for bipedal robots with input constraints and underactuation has the potential to enable locomotion in dynamic, complex and unstructured environments. 
Yet, the high-dimensional nature of bipedal robots has limited the use of full-order rigid body dynamics to
gaits which are synthesized offline and then tracked online. In this work we develop an online nonlinear model predictive control approach that leverages the full-order dynamics to realize diverse walking behaviors. Additionally, this approach can be coupled with gaits synthesized offline via a desired reference to enable a shorter prediction horizon and rapid online re-planning, bridging the gap between online reactive control and offline gait planning. We demonstrate the proposed method, both with and without an offline gait, on the planar robot AMBER-3M in simulation and on hardware.

\end{abstract}


\section{Introduction} \label{sec:intro}



From complex terrains inaccessible by wheels to human-centered infrastructure impractical for quadrupeds, bipedal robots hold the potential to operate in diverse environments in which other robots struggle. To achieve this potential, it is necessary to demonstrate a rich set of locomotion behaviors that are dynamically stable. Bipedal robots capable of demonstrating diverse behaviors, much like their human counterparts, leverage phases of underactuation. This underactuation necessitates the dynamic coordination of the whole-body dynamics of the robot---planning for the next foot strike must occur throughout the step---in a manner that accounts for the inherently nonlinear passive dynamics of the system. Achieving diverse locomotion behaviors in complex environments, therefore, motivates that whole-body planning be done on the robot in real-time, thereby going beyond pre-planned periodic walking gaits.    




The challenge of underactuation present in bipedal locomotion has historically been approached through the synthesis of gaits, i.e. dynamically stable reference trajectories. Many of these approaches for gait synthesis use condensed stability conditions like Zero-Moment Point (ZMP) \cite{vukobratovic2004zero}, or rely on other reduced-order models that simplify elements such as leg mass \cite{kajita20013d, koolen2012capturability, wensing2013high}. Alternatively, the method of Hybrid Zero Dynamics (HZD) has presented a tool for synthesizing \emph{periodic gaits} that account for the underactuated and hybrid nature of the full system dynamics \cite{westervelt2007feedback}. 
Not only do gaits synthesized via HZD possess formal stability guarantees, but they have shown great efficacy when deployed experimentally \cite{sreenath2011compliant, reher2019dynamic}.
Despite these successes, ensuring stability guarantees for high-dimensional bipedal systems often induces computational requirements that limit gait synthesis via HZD to an offline procedure. Adding a measure of flexibility to bipedal locomotion is often done by synthesizing a library of gaits \cite{gong2019feedback, nguyen2020dynamic, reher2021inverse}, although this is limited to periodic gaits, requires complex engineering solutions for gait transitions and ensuring constraint satisfaction during execution.



\begin{figure}[t]
    \centering
    \includegraphics[width=0.79\linewidth]{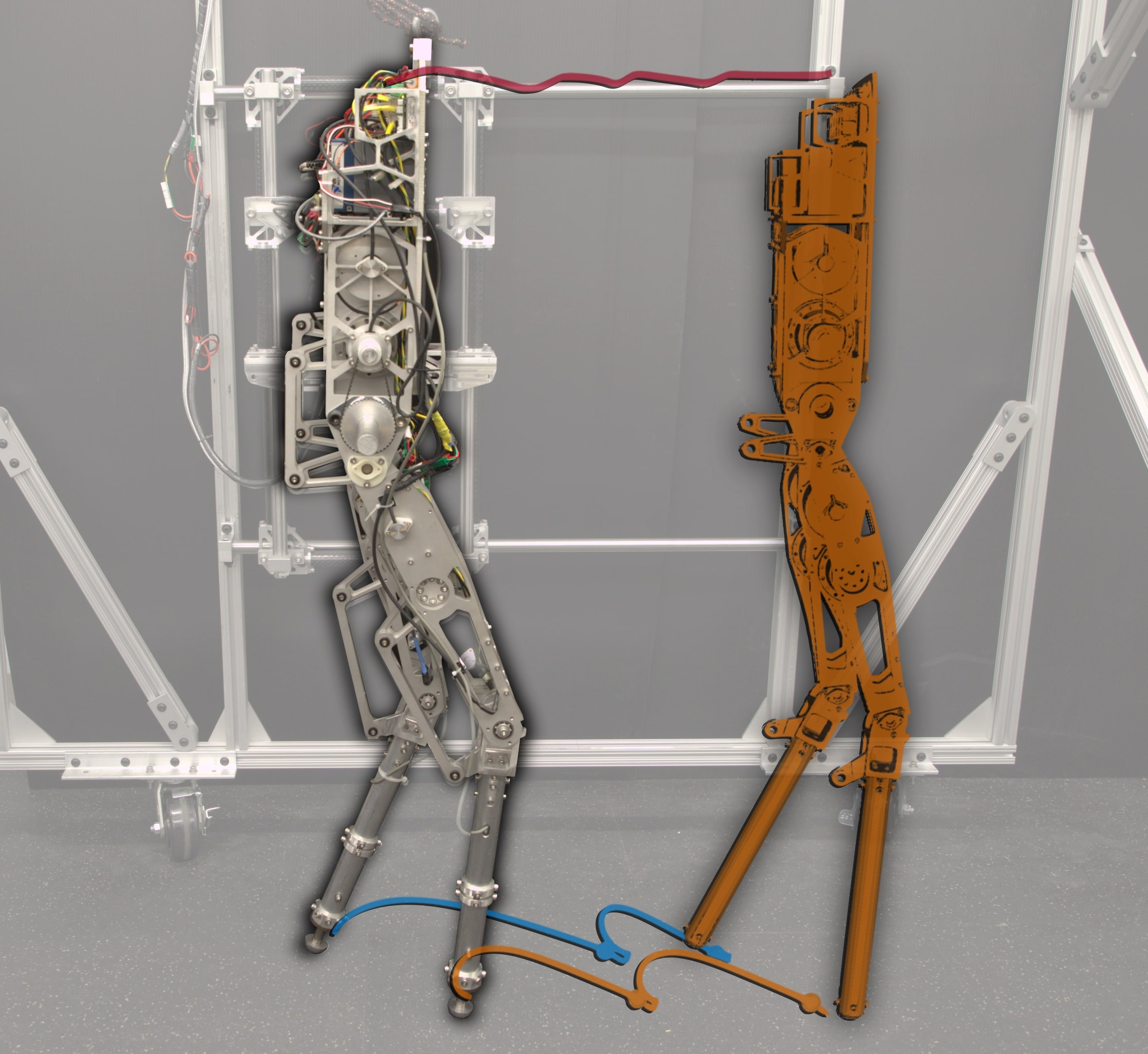}
    \caption{AMBER-3M platform using the whole-body nonlinear MPC incorporating an HZD gait. The optimized feet and torso trajectories are visualized along the prediction horizon.}
    \label{fig:opener}
    \vspace{0mm}
\end{figure}

\begin{figure*}
    \centering
    \includegraphics[width=0.95\textwidth]{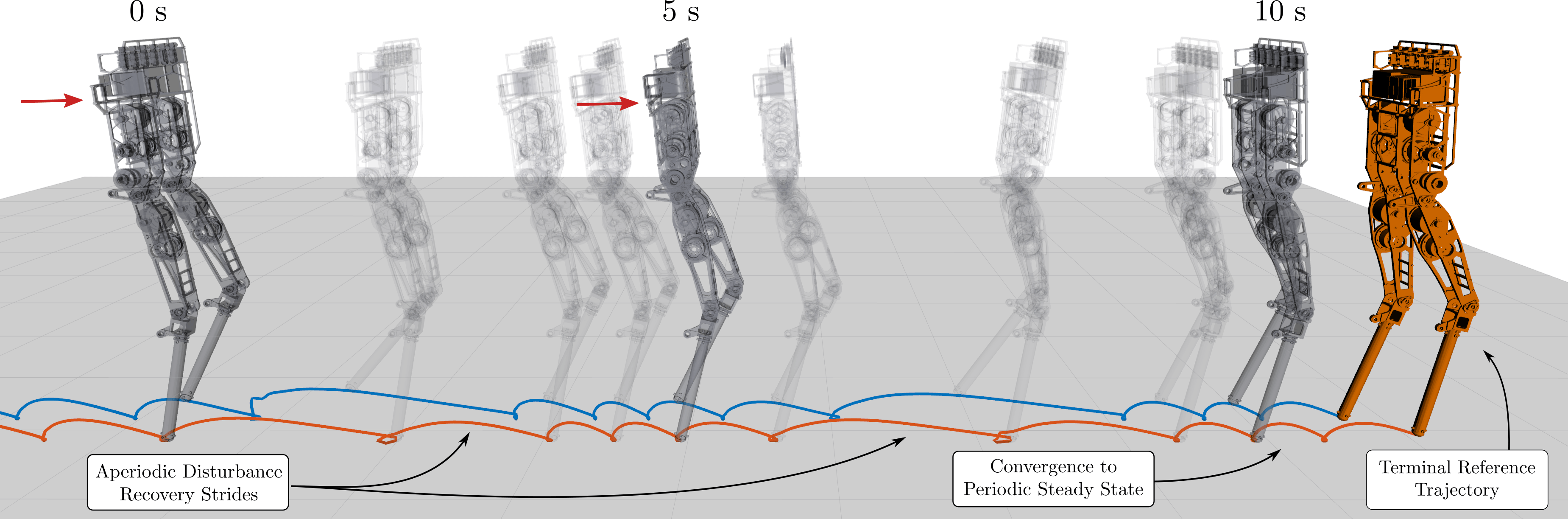}
    \caption{Push recovery using the proposed method under a disturbance -- notice the aperiodic stepping that was planned online in order to reject the disturbance, something that is not possible with traditional HZD based methods.\vspace{-4.9mm}}
    \label{fig:push_recovery}
\end{figure*}

In contrast, Model Predictive Control (MPC) provides a tool for the online synthesis of general, \emph{aperiodic trajectories}, allowing feedback of environmental parameters to be incorporated into dynamic motion planning \cite{mastalli2020crocoddyl,sleiman2021unified}. In particular, by optimizing directly over contact forces, these methods have seen significant use in online motion planning for quadrupedal robotics, with extensive experimental results \cite{dicarlo2018dynamic,grandia2019feedback, villarreal2020mpc}.
Online motion planning for bipedal robots has typically required ZMP conditions \cite{krause2012stabilization, tedrake2015closed, scianca2020mpc}, used simplified template models for planning and whole-body tracking controllers \cite{naveau2016reactive, kuindersma2016optimization, apgar2018fast, xiong20223d}, or used whole-body planning only for non-walking tasks such as reaching \cite{koenemann2015whole}. Whole-body motion planning results for bipedal walking have been predominantly in simulation \cite{diedam2008online, faraji2014versatile, dai2014whole, brasseur2015robust}, or cancelled nonlinear dynamics through feedback-linearization before planning \cite{powell2015model}. Notably, the online motion planning tools with remarkable experimental results for quadrupedal locomotion have not yet achieved commensurate results for bipedal robotics.


One of the key challenges in online whole-body motion planning is computational limitations, as producing stable locomotion requires optimizing over a sufficiently long horizon.
The methods for quadrupeds that have yielded experimental results typically exploit low leg inertia to neglect leg dynamics, reducing the state dimension in the optimization. 
%
Transferring this reduction to bipedal systems is difficult, however, as the legs compose a relatively high fraction of the system's total inertia.
At the same time, simultaneously considering both leg and torso dynamics results in many degrees of freedom, making optimization over long time horizons computationally intensive. 
Furthermore, the narrow stance width and high center of mass of bipeds necessitate a high planning frequency to counteract disturbances in underactuated dimensions.
Thus, it is paramount to design whole-body motion planners that balance the trade-off between horizon length, model complexity, and planning frequency.

%




We make three contributions in this work. First, we propose a nonlinear MPC approach for online whole-body motion planning of bipedal robotic locomotion based on existing methods used for quadrupedal locomotion \cite{farshidian2017efficient}, which achieved a wide range of stable behaviors with a planning horizon of \SI{2}{\second} and update frequency of up to \SI{270}{\Hz}. Second, to reduce the computational burden of online whole-body planning, we incorporate a stable walking gait synthesized offline via Hybrid Zero Dynamics (HZD) into the nonlinear optimization problem. This information permits robust walking while optimizing over shorter horizon lengths (\SI{0.2}{\second}) that require less computational effort and allow rapid re-planning (\SI{850}{\Hz}) -- this will be important for achieving whole-body planning on 3D walking systems. 
Lastly, we experimentally validate the proposed approach on the planar bipedal robot AMBER-3M \cite{ambrose2017toward}, demonstrating standing, stepping in place, and walking. To the best of our knowledge, this is the first experimental demonstration of online whole-body motion planning for bipedal walking.

\section{Background} \label{sec:opt}

\subsection{System Dynamics}

As walking consists of phases of intermittent contact with the world, it is naturally modeled as a hybrid system with of phases of continuous dynamics followed by discrete transition events. The configuration of the robot may be described by a set of $d$ generalized coordinates:
\begin{equation}
\vspace{.5mm}
\q = \begin{bmatrix}\q_b^\top & \q_j^\top \end{bmatrix}^\top\in\mathcal{Q} \triangleq SE(3) \times \mathcal{Q}_j,
\vspace{.5mm}
\end{equation}
which include the base coordinates $\q_b$ and joint coordinates $\q_j$ of the robot, respectively. To capture the various contact modes the robot may evolve under,
consider a collection of domains $\{\mathcal{D}_c\}$ with $\mathcal{D}_c\subseteq \mathcal{X}$ for $c=1,\ldots, p$, where $\mathcal{X}$ is the tangent bundle of $\Q$ and $p$ denotes the number of contact modes. Associated with these domains are a collection of guards $\{\mathcal{S}_c\}$ with $\mathcal{S}_c\subseteq \mathcal{X}$ and reset maps $\{\Delta_c\}$ with $\Delta_c:\mathcal{X}\to \mathcal{X}$ which are used to define how the system behaves during transitions between contact modes. Additionally, in each domain $\mathcal{D}_c$ the coordinates of the robot are subject to a collection of $n_c$ \emph{holonomic constraints} with associated contact Jacobians $\mb{J}_c:\Q\to\R^{6 n_c \times d}$.

Using the Euler-Lagrange method, the system dynamics in a given domain $\mathcal{D}_c$ are given by:
\begin{align}
\vspace{.5mm}
    \mb{D}(\q)\ddot{\q}+\mb{h}(\q, \dot \q) &= \mb{B}(\q)\bs{\tau} +  \mb{J}_{c}(\q)^\top\bs{\lambda},
    \label{eq:full_rigid_body_dynamics} \\
    \mb{J}_c(\q) \ddot{\q} + \dot{\mb{J}}_c(\q,\dot{\q}) \dot{\q} &= \mb 0, \label{eq:jacobian_dyn_constraint}
    \vspace{.5mm}
\end{align}
with symmetric positive definite inertia matrix  $\mb{D}:\Q\to\mathbb{S}_{\succ 0}^d$, centrifugal, Coriolis, and gravitational terms $\mb{h}:\mathcal{X}\to\R^{d}$, actuation matrix $\mb{B}:\Q\to\R^{d\times m}$, torques $\bs{\tau}\in\R^m$, and constraint forces $\bs{\lambda}\in\R^{6n_c}$. Defining the state $\x\in \mathcal{X}$ as:
\begin{equation}
\vspace{.5mm}
\x=\begin{bmatrix}\q^\top&\dot{\q}^\top\end{bmatrix}^\top,
\label{eq:mpc_state}
\vspace{.5mm}
\end{equation}
and solving for the constraint forces $\bs{\lambda}$ via \eqref{eq:full_rigid_body_dynamics}-\eqref{eq:jacobian_dyn_constraint}, the system dynamics in a given domain $\mathcal{D}_c$ can be rewritten as:
\begin{align}
\label{eqn:control_affine}
\vspace{.5mm}
    \dot{\mb{x}} =  \underbrace{\begin{bmatrix}\ \dot \q \\ \mb{D}^{-1}(-\mb{h}+ \mb J_c^\top \bs{\lambda})\end{bmatrix}}_{\mb f_c(\x)} + \underbrace{\begin{bmatrix}\mb 0\\\mb {D}^{-1}\mb B \end{bmatrix}}_{\mb g(\x)}\bs \tau,
\end{align}
where the dependence on $\q$ and $\dot{\q}$ has been dropped for notational simplicity. The resulting functions $\mb{f}_c:\mathcal{X}\to\mathbb{R}^{2d}$ and $\mb{g}:\mathcal{X}\to\mathbb{R}^{2d\times m}$ are assumed to be continuously differentiable on the domain $\mathcal{D}_c$.

To model a transition from contact mode $c$ to contact mode $c'$, consider a state $\x^-\triangleq(\q^{-},\dot{\q}^-)\in\mathcal{S}_c$. The discrete transition map is given by:
\begin{align}
\mb D(\q^-)(\dot \q^+ - \dot \q^-) &= \mb J_{c'}^\top(\q^-) \mb F, \\ 
 \mb J_{c'}(\q^-) \dot {\q}^+ &= \mb 0,
\end{align}
whereby solving for the impulse force $\mb F\in\R^{n_c}$ yields:
\begin{equation}
    \x^+ \triangleq \begin{bmatrix} \q^+ \\ \dot{\q}^+ \end{bmatrix} =  \underbrace{\begin{bmatrix} \q^- \\ \dot{\q}^- + \mb D(\q^-)^{-1}\mb J_{c'}^\top(\q^-) \mb F\end{bmatrix}}_{\Delta_c(\x^-)}.
    \label{eq:impulse_dynamics}
\end{equation}

\subsection{Nonlinear Model Predictive Control}
Nonlinear MPC solves a optimization problem in a receding horizon manner by solving the following finite time nonlinear optimal control problem:
\begin{subequations}
\begin{align}
     & \underset{\mb u(\cdot)}{\text{minimize}} && \phi(\mb x(t_{H})) + \int_0^{t_{H}} l(\mb x(t), \mb u(t), t)\mathrm{d}t, \label{eq:mpc_cost} \\
    &\text{subject to:} && \mb x(0) = \mb x_0, 
    \label{eq:mpc_initial} \\
    & & &  \dot{\mb x} =  \mb f(\mb \x) + \mb g(\mb \x) \mb u,  \label{eq:mpc_dynamics} \\
    & & & \mb{x}(t^+_{i}) = \Delta_c(\mb{x}(t_{i})),
    \label{eq:mpc_jumpmap} \\
    & & & \mb h_{eq}(\mb x, \mb u, t) =  \mathbf {0},  \label{eq:mpc_eqconstraint} \\
    & & & \mb h_{in}(\mb x,\mb u, t) \geq  \mathbf {0} \label{eq:mpc_inequality},
\end{align}
\label{eq:mpc_formulation}%
\end{subequations}
where $t_{H}$ is the length of the horizon, $\phi:\mathcal{X}\to\R$ is the terminal cost, $l:\mathcal{X}\times \R^m\times\R\to\R$ is the time-varying running state-input cost, and $t_i$ are times of contact mode transitions. The optimal control problem is solved in real-time by updating the initial conditions \eqref{eq:mpc_initial} with the measured state of the system. Eq.~\eqref{eq:mpc_dynamics} describes the system dynamics. $\mb h_{eq}:\mathcal{X}\times \R^m\times\R\to\R^{eq}$ and $\mb h_{in}:\mathcal{X}\times \R^m\times\R\to\R^{in}$ are generalized path equality and inequality constraints, respectively. There exist various approaches to solve this problem as outlined in \cite{rawlings2017model}. We take a direct-multiple shooting transcription of the problem together with a sequential quadratic programming approach to handle nonlinearities~\cite{diehl2006fast}. Inequality constraints \eqref{eq:mpc_inequality} are implemented  through relaxed-barrier penalty functions \cite{feller2016relaxed}.


A key component in establishing closed loop stability and recursive feasibility is the choice of terminal components~\cite{mayne2000constrained}, either as terminal cost in~\eqref{eq:mpc_cost} or as constraints on the terminal state, $\mb x(t_{H})$, to lie in a control invariant set. In practice, for nonlinear complex systems, it is challenging to prove that such conditions hold. Extending the prediction horizon is a common choice to reduce the relative importance of the terminal components~\cite{mayne2014model}. However, for systems where long prediction horizons are not feasible due to computational limits, careful choice of terminal components directly translates to the overall performance of the controller, as we will empirically show in this work.

\begin{figure*}[tb!]
    \centering
    \includegraphics[width=0.95\linewidth]{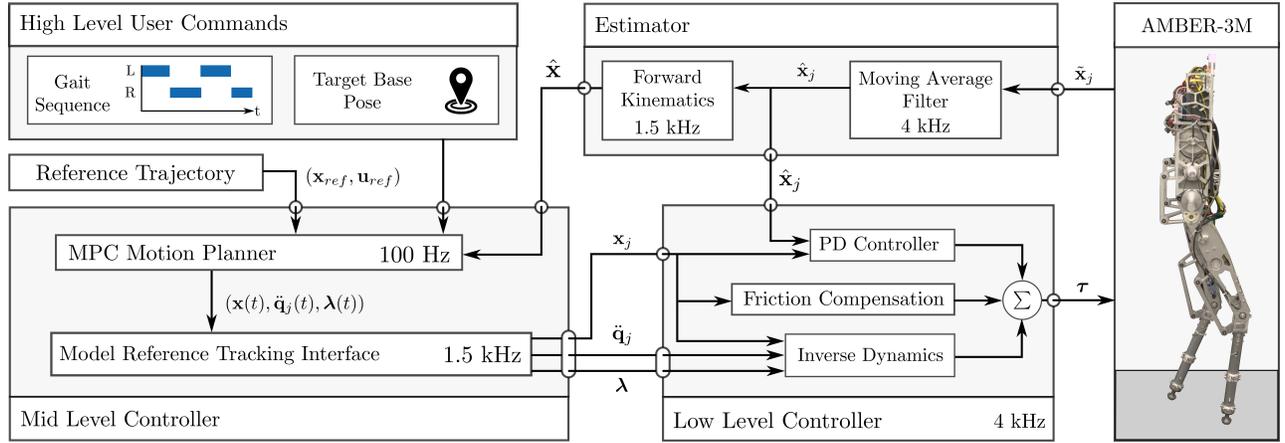}
    \caption{Multi-rate control architecture incorporating whole-body planning via MPC and low-level tracking controller.}
    \vspace{-4mm}
    \label{fig:toy_example}
\end{figure*}
\subsection{Hybrid Zero Dynamics (HZD)}
The HZD framework provides a formal method for producing walking behaviors for robotic systems, and has been successfully employed on a variety of platforms \cite{westervelt2007feedback,sreenath2011compliant, reher2019dynamic}. Synthesis of gaits via HZD is centered around defining \emph{outputs} $\mb y:\mathbb{R}^d\times\R^{r}\to \mathbb{R}^o$ as:
\begin{align}
    \mb y(\q,\bs{\alpha}) = \mb y_a(\q) - \mb y_d(\q,\bs{\alpha}),
\end{align}
where $\mb y_a:\mathbb{R}^d\to \mathbb{R}^o$ and $\mb y_d:\mathbb{R}^d\times\R^r\to \mathbb{R}^o$ are the \emph{actual} and \emph{desired} outputs, respectively, and are assumed to be continuously differentiable. The actual outputs $\mb{y}_a$ are chosen to satisfy a controllability property allowing them to be driven to the desired outputs \cite{westervelt2007feedback}. The desired outputs $\mb{y}_d$ depend on the set of parameters $\bs{\alpha}\in\R^r$, which are chosen to regulate the underactuated coordinates of the system. More precisely, the \emph{zero dynamics manifold} is defined as the subspace of state coordinates for which the outputs and their derivatives are zero:
\begin{align}
    \mathcal{Z}_c = \{(\q,\dot \q) \in \mathcal{X} : \mb y(\q,\bs{\alpha}) = \dot{\mb y}(\q, \dot \q,\bs{\alpha}) = \mb 0 \}.
\end{align}
The parameters are then chosen to satisfy the \emph{hybrid invariance condition}, $
\Delta_c(\mathcal{Z}_c \cap \mathcal{S}_c) \subset \mathcal{Z}_c$, for each contact mode $c$, which ensures that the underactuated coordinates of the system remain stable through impacts. Ensuring this condition is achieved by finding values of $\bs{\alpha}$ through  nonlinear optimization \cite{hereid2017frost}. Given a desired trajectory $\mathbf{y}_d$ from an HZD optimization program, a time-varying state and input trajectory $(\mb x_{\rm ref}(t), \mb u_{\rm ref}(t))$ can be reconstructed. Importantly, as this trajectory is designed for the full-order hybrid dynamics, it will serve as a control invariant that will be incorporated in our MPC formulation. 


 \section{Whole-Body Motion Planning \& Control} \label{sec:framework}

Our nonlinear MPC problem will be constructed using the \emph{OCS2} toolbox~\cite{OCS2}, which provides convenient interfaces to the \emph{Pinocchio}~\cite{carpentier2019pinocchio} rigid body library and \emph{CppAd}~\cite{bell2012cppad} automatic differentiation tools. 
Our formulation assumes that the contact schedule associated with a given locomotion mode (standing, stepping in place, walking) is given by the user. 
The fixed contact schedule assumption simplifies the optimization problem as the sequence of domains and timing of contact mode transitions does not need to be optimized \cite{farshidian2017efficient,mastalli2020crocoddyl}. The position of the foot at contact is captured in the optimization problem through its kinematic relationship with joint coordinates. Moreover, we assume the user provides a desired base pose and velocity to the MPC. In this section we discuss the formulation of bipedal locomotion planning as an MPC problem as posed in~\eqref{eq:mpc_formulation}.

%

\subsection{System Dynamics}

Due to the affine relationship between generalized accelerations $\ddot{\mb{q}}$, torques $\bs{\tau}$, and contact forces $\bs{\lambda}$ in~\eqref{eq:full_rigid_body_dynamics}, and assuming the torques do not directly impact the floating-base equations of motion, the system dynamics in~\eqref{eqn:control_affine} may be rewritten to interpret the joint accelerations $\ddot{\mb{q}}_j$ and contact forces $\bs{\lambda}$, instead of the torques $\bs{\tau}$, as inputs. The computational benefit of this reparametrization has been shown for reactive whole-body control~\cite{bellicoso2017dynamic} and offline trajectory optimization~\cite{ferrolho2021inverse}. To see this, we write the dynamics \eqref{eq:full_rigid_body_dynamics} in terms of non-actuated base coordinates and fully actuated joint coordinates: 
\begin{equation}
\begin{bmatrix} \mb D_{bb}, & \mb D_{bj} \\ \mb D_{bj}^\top & \mb D_{jj} \end{bmatrix}  \begin{bmatrix} \ddot{\mb{q}}_b \\ \ddot{\mb{q}}_j \end{bmatrix} +  \begin{bmatrix} \mb{h}_b \\ \mb{h}_j \end{bmatrix} = \begin{bmatrix} \bm{0} \\ \mb B_j \end{bmatrix}\bs{\tau} + \begin{bmatrix} \mb J_{c,b}^\top \\  \mb J_{c,j}^\top \end{bmatrix} \bs{\lambda}.
\end{equation}
The base acceleration may be expressed as:
\begin{equation}
\ddot{\mb{q}}_b = -\mb{D}_{bb}^{-1}\left(\mb{h}_b + \begin{bmatrix} \mb{D}_{bj} &  -\mb{J}_{c,b}^\top \end{bmatrix}  \begin{bmatrix} \ddot{\mb{q}}_j \\  \bs{\lambda} \end{bmatrix}\right),
\label{eq:reparam_base}
\end{equation}
and assuming the legs are fully actuated ($\mb{B}_j$ is invertible), the corresponding joint torques may be expressed as:
\begin{equation}
\bs \tau = \mb{B}_j^{-1}\left(\mb{D}_{bj}^\top \ddot{\mb{q}}_b  + \mb{h}_j +  \begin{bmatrix} \mb D_{jj} &  -\mb{J}_{c,j}^\top  \end{bmatrix}  \begin{bmatrix} \ddot{\mb{q}}_j \\  \bs{\lambda} \end{bmatrix}\right),
\label{eq:reparam_torque}
\end{equation}
maintaining an affine dependence on $\ddot{\mb{q}}_j$ and $\bs{\lambda}$. The base dynamics in~\eqref{eq:reparam_base} fully encode the challenge of under-actuation and encapsulate the core of the floating-base dynamics. Equation~\eqref{eq:reparam_torque} plays a secondary role and is only required when formulating torques constraints. We may view the control inputs to optimize over as:
\begin{equation}
\mb{u} = \begin{bmatrix}  \ddot{\mb{q}}_j^\top, & \bs{\lambda}^\top \end{bmatrix}^\top,
\label{eq:mpc_input}
\end{equation}
with the corresponding system dynamics defined as:
 \begin{equation}
\dot{\mb{x}} = \begin{bmatrix} \dot{\mb{q}} \\  \mb{D}_{bb}^{-1}\left(-\mb{h}_b - \mb{D}_{bj}\ddot{\mb{q}}_j + \mb{J}_{c,b}^\top \bs{\lambda} \right) \\ \ddot{\mb{q}}_j \end{bmatrix}.
\label{eq:reparameterized_dynamics}
 \end{equation}

In order to avoid large discontinuities in the optimized trajectory, the contact transition maps in~\eqref{eq:mpc_jumpmap} have been set to identity maps for the online MPC program, with exponential damping of the contact point velocity after impact being regulated through the stance foot constraint in~\eqref{eq:stancefoot_constraint}, defined in section~\ref{sect:constraints}. Inclusion of the impact dynamics in~\eqref{eq:impulse_dynamics} will be pursued in future work. Due to the assumption of a fixed contact schedule, inclusion of the contact transition map does not fundamentally change the complexity of the optimization problem \cite{mastalli2020crocoddyl}.

\subsection{Cost Functions}
\label{sec:cost}
The cost function is formulated as a nonlinear least square cost around a given state and input reference trajectory. To that end we define the set of tracking errors as follows:
 \begin{equation}
 \bs{\epsilon}_{x} = \mb{x} - \mb{x}_{\rm ref}(t), \,
 \bs{\epsilon}_{u} = \mb{u} - \mb{u}_{\rm ref}(t), \,
   \bs{\epsilon}_{i} = \begin{bmatrix}
   \mb{p}_{i}  - \mb{p}_{i,\rm ref}(t)  \\ 
   \mb{v}_{i} - \mb{v}_{i,\rm ref}(t)  \\ 
   \mb{a}_{i} - \mb{a}_{i,\rm ref}(t), 
   \end{bmatrix},\notag
 \end{equation}
where $\mb{x}_{\rm ref}$ is the state reference, $\mb{u}_{\rm ref}$ is the input reference, and $\mb{p}_{i},\mb{v}_{i},\mb{a}_{i}\in\R^3$ with $i\in\{1,2\}$ are the Cartesian position, velocity, and accelerations of the $i^{th}$ foot, with corresponding references $\mb{p}_{i,\rm ref},\mb{v}_{i,\rm ref},\mb{a}_{i, \rm ref}$. The references $\mb{x}_{\rm ref}$ and $\mb{u}_{\rm ref}$ are defined heuristically (see Section \ref{sec:referencetrajs}) or via a walking gait synthesized offline using HZD. The running state-input cost $l$ is given by:
\begin{equation}
  l(\mb{x}, \mb{u}, t) = \frac{1}{2}\bs{\epsilon}_{x}^\top\mb{Q}\bs{\epsilon}_{x}
  + \frac{1}{2}\bs{\epsilon}_{u}^\top\mb{R}\bs{\epsilon}_{u}
  + \frac{1}{2}\sum_{i} \bs{\epsilon}_{i}^\top \mb{W}\bs{\epsilon}_{i} ,
  \label{eq:intermediate_cost}
\end{equation}
where $\mb{Q},\mb{R}$, and $\mb{W}$ are positive definite weighting matrices.
 
To pick an appropriate weighting for the terminal cost, we approximate the infinite horizon cost by solving an unconstrained Linear Quadratic Regulator (LQR) problem using a linear approximation of the dynamics and a quadratic approximation of the running costs~\eqref{eq:intermediate_cost} around the nominal stance configuration of the robot. The positive definite Riccati matrix $\mb{S}_{\text{LQR}}$ of the cost-to-go is used to define the quadratic cost around the terminal reference state:
\begin{equation}
\phi(\mb{x}) = \frac{\rho}{2} \bs{\epsilon}_{x}(T)^\top \mb{S}_{\text{LQR}} \bs{\epsilon}_{x}(T),
\end{equation}
where $\rho > 0$ is a hyperparameter. Setting $\rho = 1.0$ would express approximately equal importance of the integrated running cost and terminal cost, and $\rho \rightarrow \infty$ would make the terminal cost behave as an equality constraint. We found good performance for the heuristic reference at $\rho = 1.0$ and for the HZD reference at $\rho = 10.0$. Note that this cost does not penalize deviation from the stance configuration used to produce the LQR solution, but rather provides a systematic way to scale the relative importance of all of the degrees of freedom of the robot in the cost.

\subsection{Reference Trajectories}
\label{sec:referencetrajs}

\paragraph{HZD Trajectory} 
HZD state and input reference trajectories, $\mb{x}_{\rm ref}(t)$ and $\mb{u}_{\rm ref}(t)$, are found offline for the whole-body nonlinear dynamics using the FROST toolbox \cite{hereid2017frost} and stored as B\'ezier polynomials. This process is completed by fixing a target gait sequence and a forward velocity, and adding various other state and input constraints to a nonlinear trajectory optimization program which ensure the underactuated dynamics of the system display stable periodic behavior. For planar systems, stability can be enforced directly in the optimization program \cite{ames2014human}, and for general systems it can be verified \textit{a-posteriori} via the Poincar\'e return map \cite{westervelt2007feedback}. The foot references $\mb{p}_{i,\rm{ref}}(t)$, $\mb{v}_{i,\rm{ref}}(t)$, and $\mb{a}_{i,\rm{ref}}(t)$ are entirely determined by $\mb{x}_{\rm ref}(t)$.

\begin{figure}[t]
    \centering
    \includegraphics[width=1.0\linewidth]{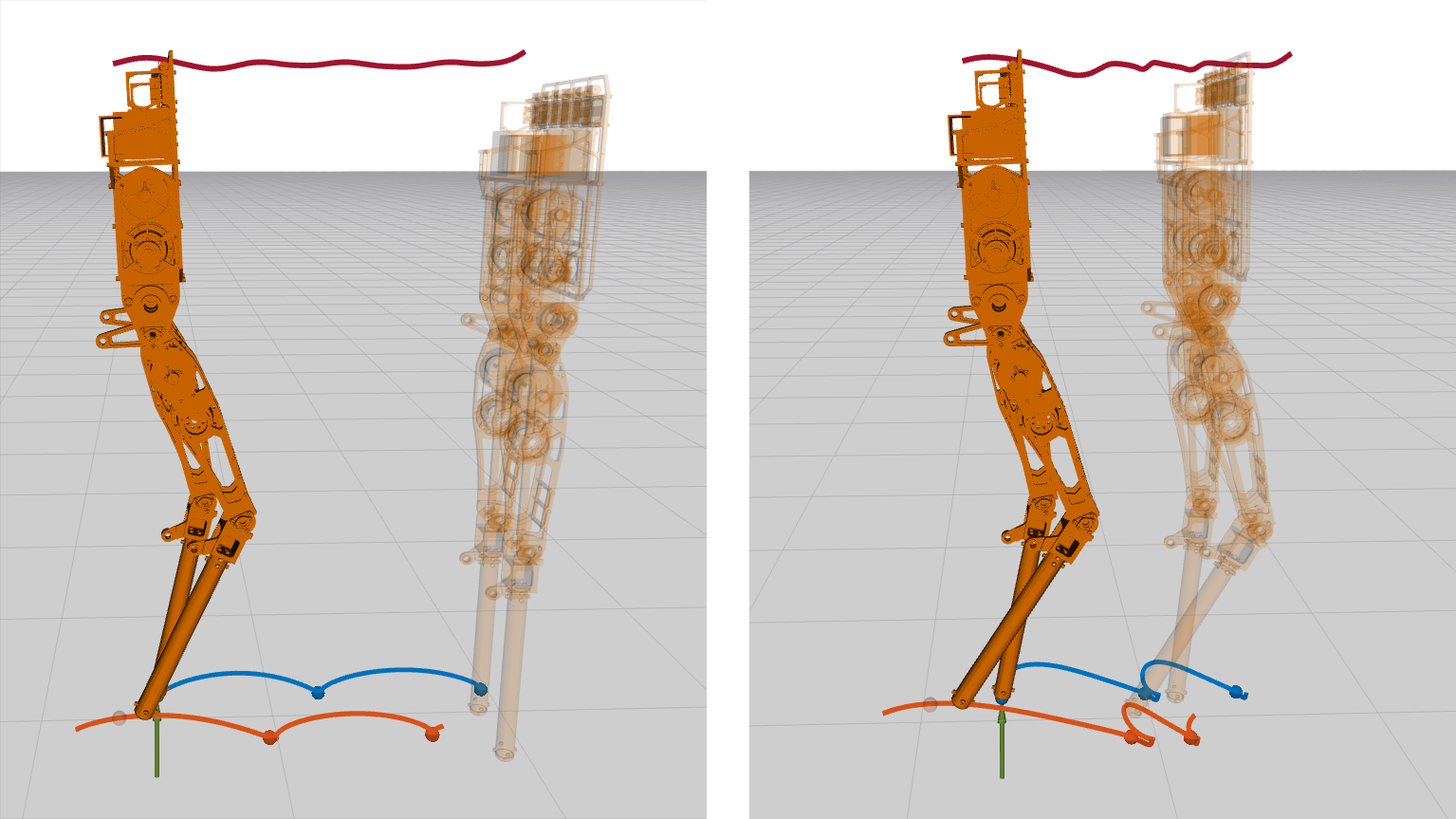}
    \caption{Heuristic (left) and HZD (right) terminal states.}
    \label{fig:terminal_state}
\end{figure}

\paragraph{Heuristic Trajectory} 
\label{sec:heuristic_reference}
To evaluate the relative impact of using a gait synthesized offline via HZD in the cost function, we 
produce a heuristic reference trajectory to be compared against. In particular, the state trajectory $\mb{x}_{\rm ref}(t)$ is composed of a user-commanded base pose and velocity, and a static nominal joint configuration. The input reference $\mb{u}_{\rm ref}(t)$ is defined with zero joint accelerations and contact forces that are evenly distributed among each foot in contact in the nominal joint configuration such that the weight of the robot is compensated. The foot references $\mb{p}_{i,\rm{ref}}(t)$, $\mb{v}_{i,\rm{ref}}(t)$, and $\mb{a}_{i,\rm{ref}}(t)$ are designed by extracting the nominal touchdown and liftoff locations below the hip at the middle of the contact phase and fitting a smooth hand-designed swing reference trajectory. The heuristic and HZD-based terminal state are visualized in Fig.~\ref{fig:terminal_state}.

\subsection{Constraints}
\label{sect:constraints}
The following constraints are imposed in problem \eqref{eq:mpc_formulation}. All inequality constraints are implemented as soft constraints with relaxed log barrier functions \cite{OCS2}.

\paragraph{Gait-Dependant Constraints} These constraints capture the different modes of each leg at any given point in time determined by the specified gait sequence. We enforce the user-defined gait and avoid foot scuffing of a swing leg by constraining the swing foot motion in the orthogonal direction to the ground surface, $\mb n\in\R^3$, to follow the Cartesian reference trajectory:
\begin{align}
\mb n^\top (\mb{a}_{i} - \mb{a}_{i, \rm ref}(t) & +    k_d (\mb{v}_{i} - \mb{v}_{i, \rm ref}(t)) \nonumber \\ & +
k_p (\mb{p}_{i} - \mb{p}_{i,\rm ref}(t))) = \bm{0}
\end{align}
where $k_d,k_p\in\R_{\geq 0}$ are feedback gains chosen to achieve asymptotic tracking in the constrained space. The foot position in the direction parallel to the ground is not directly constrained; rather, tracking is enforced via the cost function described in Section \ref{sec:cost}. For a stance leg we enforce a stationarity constraint in Cartesian space through:
\begin{equation}
\mb a_{i} + k_d \mb v_{i} = \mathbf{0}
\label{eq:stancefoot_constraint}
\end{equation}

\paragraph{Contact Force Constraints}
The following constraints require the contact forces at each foot to match the designation of swing and stance legs: %
\begin{align}
&\left\{ 
\begin{array}{ll}
		\bm \lambda_{i}  = \mb 0, \qquad & i\textrm{ is a swing leg} \\
		\bm \lambda_{i} \in \mathcal{C}(\mb n, \mu_c), \qquad & i\textrm{ is a stance leg.}
\end{array}
\right.
\end{align}
The first constraint requires no contact force from a swing leg, as it does not contact the ground. The second constraint requires the contact force of a stance leg to lie in the friction cone $\mathcal{C}(\mb{n},\mu_c)$ defined by the surface normal $\mb{n}$ and the friction coefficient ${\mu_c = 0.6}$. This is a second-order cone constraint and is expressed in the local surface aligned frame:
\begin{equation}
    \mu_c \lambda_{i,3} - \sqrt{\lambda_{i,1}^2 + \lambda_{i,2}^2} \geq 0.
    \label{eq:cone}
\end{equation}
Note that only linear forces are present as the robot has point feet, and that the friction constraint also enforces a unilateral contact constraint as it requires $\lambda_{i,3} \geq 0$.

\paragraph{Joint and Torque Limits} \label{paragraph:limits}
The joint coordinates and joint coordinate velocities are enforced to lie in the set of minimum and maximum joint positions and velocities through state inequality constraints: $\mb{x} \in [ \mb{x}_{\text{min}}, \mb{x}_{\text{max}}]$. Similarly, the joint torques can be computed by Eq.~\eqref{eq:reparam_torque} and should lie within joint torque limits $\bs{\tau}  \in [ \bs{\tau}_{\text{min}}, \bs{\tau}_{\text{max}}]$.

\subsection{Low-Level Controller}
As shown in Fig.~\ref{fig:toy_example}, the state and input trajectories generated by MPC are interpolated at a high frequency and converted to a feed-forward control torques, ${\bm \tau}_{\text{MPC}}$, via~\eqref{eq:reparam_torque}.
%
%
As the feed-forward torque is model-based, we compensate for model errors when executing the controller on hardware by adding a proportional-derivative torque, ${\bm \tau}_{\text{PD}}$, and a friction compensation torque, ${\bm \tau}_{\text{FC}}$, to the feed-forward torque:
\begin{equation}
    {\bm \tau} = {\bm \tau}_{\text{MPC}} + {\bm \tau}_{\text{PD}} + {\bm \tau}_{\text{FC}}.
    \label{torque_limits}
\end{equation}

The total torque $\bm \tau$ is send to the open loop torque controlled BLDC motors.







\section{AMBER Implementation \& Results} 
The AMBER-3M platform is a 5-link planar bipedal robot, which has four open loop torque controlled BLDC motors connected via harmonic drives to the hip and knee joints. The total mass of the robot amounts to \SI{21.6}{\kilo\gram}, approximately \SI{40}{\percent} of which is located in the legs. The joint coordinates are given by $\mb q_j\in\mathcal{Q}_j\subset \R^4$, and, due to the planar nature of the robot, the base coordinates are given by $\mb{q}_b = \begin{bmatrix} x_b, z_b, \theta_b\end{bmatrix}^\top\in SE(2)$ resulting in a state vector $\mb x \in \mathbb{R}^{14}$. The input  $\mb u = (\ddot{\mb{q}}_j, \bs\lambda)\in \mathbb{R}^{8}$ contains the joint accelerations and 2D Cartesian contact forces at the point-feet. Position and velocity measurements of the four joint angles and the base angle are present and used to estimate the linear base position and velocity. No impact measurement was used during the experiments. The time discretization in the multiple shooting scheme is set to \SI{15}{\milli\second} and we allow for a maximum of 10 SQP iterations per MPC problem. All planning, control, and estimation loops were done on separate threads on an offboard Ryzen 9 5950x CPU @ \SI{3.4}{\giga\hertz}. Benchmarks of the maximum obtainable MPC frequency for different horizon lengths can be seen in Table \ref{tab:mpc_freq}. To isolate how the system's behavior depends on horizon length, all experiments were conducted with a consistent MPC frequency of \SI{100}{\hertz}.

\begin{table}[t]
\begin{center}
\caption{MPC Planning Frequency (10 SQP Iterations)}
\begin{tabular}{l c  c  c c} 
\hline
Horizon Length [s] & 2.0 & 1.0 & 0.5 & 0.2  \\ 
 \hline
   MPC Frequency [Hz] & 270 & 480 & 670 & 850  \\ 
 \hline
\end{tabular}
\vspace{-2mm}
\label{tab:mpc_freq}
\end{center}
\end{table}

As can be seen in the supplementary video \cite{video}, the proposed MPC formulation is capable of simultaneously stabilizing the underactuated system dynamics and synthesizing valid motion trajectories for a broad range of gait pattern and target velocities both in simulation and on hardware. To evaluate the effect of changing reference signals on the feasibility and robustness of the full control pipeline, a sequence of disturbances of increasing magnitude was applied in simulation with the following MPC configurations:
\begin{itemize}
  \item \emph{MPC with No Terminal}: The proposed whole-body MPC with heuristic references for the running cost (refer to Sec.\ref{sec:referencetrajs}) and no terminal cost. 
  \item \emph{MPC with Heuristic Terminal}: Same as above, but with heuristic references included as a terminal cost.
  \item \emph{MPC with HZD Reference}: The proposed whole-body MPC with HZD-based references for the running and terminal cost (refer to Sec.~\ref{sec:referencetrajs}). 
  \item \emph{Lumped Mass MPC}: Uses a simplified dynamics model for the planning stage by moving leg inertia to the torso, otherwise identical to \emph{MPC with Heuristic Terminal}.
  \item \emph{HZD with PD}: An offline generated HZD trajectory tracked by a joint level PD controller.  
\end{itemize}
%

\begin{table}[b]
\begin{center}
\caption{Maximum disturbance rejection and step adaption range (difference between smallest and largest observed step length). MPC planning frequency clamped at 100 Hz.} 
\begin{tabular}{ l ||c |  c  | c|| c } 
 \hline 
 &\multicolumn{3}{c||}{Disturbance Rejection }& Step Range \\\hline \hline
 Horizon Length & 2 s & 0.5 s & 0.2 s& 2 s \\ \hline 
 Lumped Mass MPC & 2 N & - & - & - \\
 \hline
 MPC + No Terminal & 22 N  & - & - & 0.63 m \\ \hline 
 MPC + Heuristic & 22 N   & 22 N  & - & 0.67 m \\  \hline 
 MPC + HZD & 22 N  & 22 N  & 20 N & 1.10 m \\ 
 \hline\hline
 HZD + PD&\multicolumn{3}{c||}{30 N } & 0.14 m\\
 \hline
\end{tabular}
\label{tab:dist_rejection}
\end{center}
\end{table}

The results of these simulations are summarized in Table~\ref{tab:dist_rejection}. First, we remark that the Lumped Mass MPC model was introduced to highlight the effects of planning over the full system dynamics for the given platform. The particular structure of this model was chosen to resemble some properties of the simplified models mentioned in 
Sec.~\ref{sec:intro}, while allowing for an implementation independent comparison. Although the Lumped Mass MPC could withstand similar disturbances to the whole-body MPC for a specified standing position, it was observed to have only a marginal ability to reject disturbances during dynamic motions like stepping in place and walking, no matter the horizon length. This confirms the need for whole-body online planning methods, especially for robots like AMBER-3M which have a non-negligible mass distribution concentrated in the legs. 

\begin{figure}[t]
    \centering
    \includegraphics[width=\linewidth]{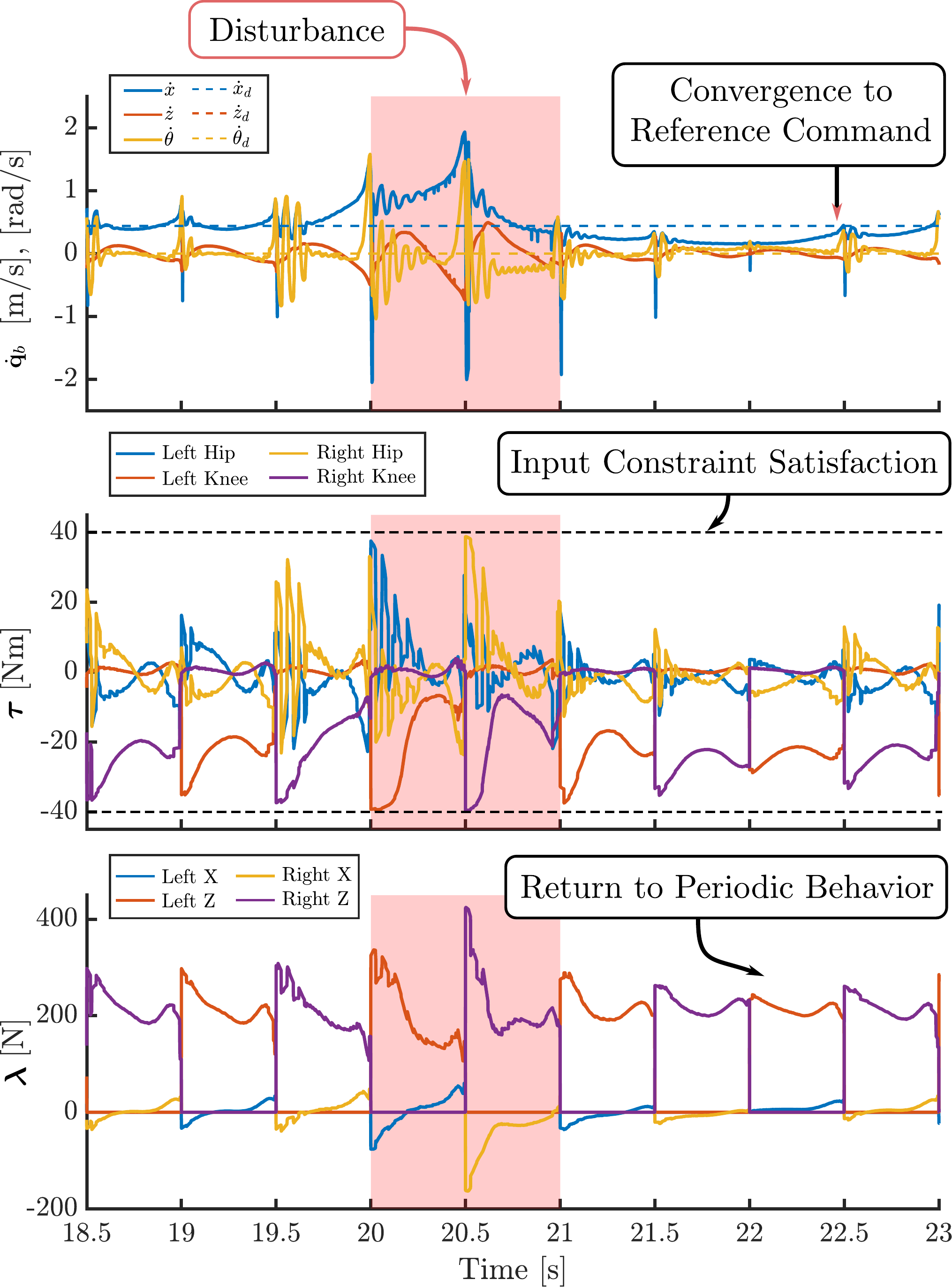}
    \caption{Simulation results for the MPC with Heuristic Terminal controller under a disturbance of \SI{20}{\newton} applied in the forward (X) direction during the marked time of \SI{1}{\second}, including states (top), torques (middle), and contact forces (bottom). The commanded forward walking velocity is \SI{0.5}{\meter/\second}. }
    \label{fig:simdata}
\end{figure}
\begin{figure*}[t]
    \centering
    \includegraphics[width=0.92\textwidth]{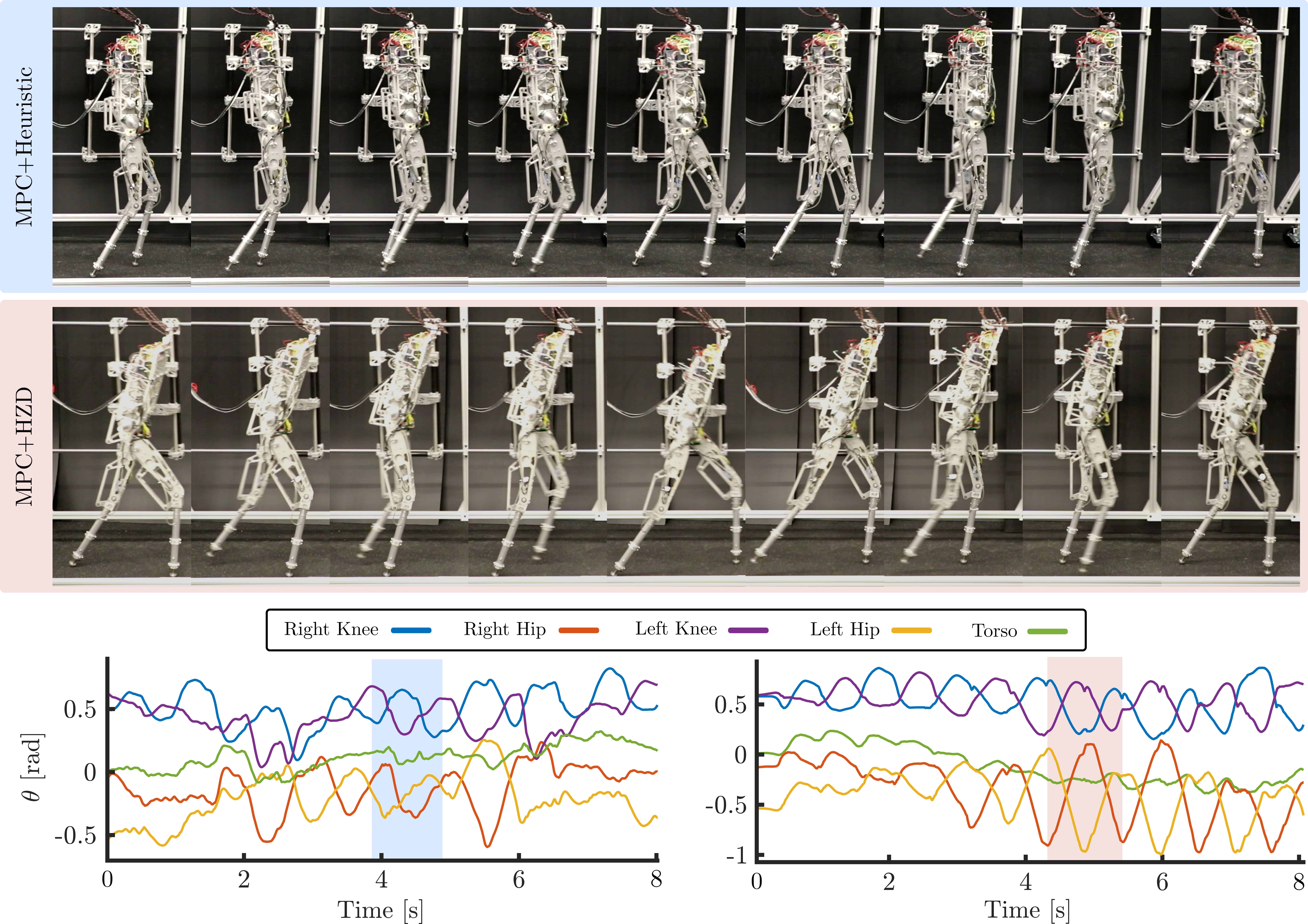}
    \caption{Gait tiles and joint angle trajectories for forward walking behavior of the whole body MPC at a horizon length of 1 second (top, left), and the whole body MPC+HZD at a horizon length of 0.5 seconds (bottom, right). The HZD reference induces stronger periodic behaviors in the joint coordinates, correlated with the periodic nature of an HZD gait.\vspace{-4.5mm}}
    \label{fig:gait_tiles}
\end{figure*}

Next, note that the MPC approach fails quickly when no terminal cost is present. When a heuristic terminal component is added, the robustness of the system dramatically increases. Furthermore, when the proposed MPC approach is combined with an HZD-based reference trajectory for running and terminal costs, the horizon length can be shortened to as low as 0.2 seconds, which drastically reduces the computational complexity. This could be essential to enable the whole body NMPC approach to be applied to a 3D biped with a larger number of degrees of freedom. These results emphasize the importance of the careful design of reference components, as their construction is tightly coupled with the performance of the overall system. Finally, it is important to note that at a disturbance of \SI{22}{\newton} during walking the foot begins to slip, causing all of the MPC based methods to fail. The HZD with PD method exhibits more robustness to foot slipping and is therefore able to endure larger disturbances, as it does not model the disturbances. 
Note that the HZD and PD method is limited to periodic motions, and during disturbance rejection it heavily restricts the allowable stepping range, as reported in Table~\ref{tab:dist_rejection}. On the other hand, the MPC methods naturally have a large variability in footstep locations in order to stabilize the system.
We believe that the ability to modulate step width and exhibit aperiodic motions will be critical for bipedal robots operating on real-world terrain. Future work will seek to combine the robustness of the HZD and PD method with the flexibility of the proposed MPC methods.


As seen in the supplementary video~\cite{video}, the various proposed approaches react differently to disturbances.
Specifically, the phase-based HZD with PD control achieves stability via implicit modification of the contact times, where the limbs are accelerated along the predefined reference trajectory. While this allows for significant disturbance rejection, it leads to the inputs being saturated for non-negligible amounts of time. On the other hand, in this MPC-based formulation, stability is achieved via explicit modification of the footstep locations, and is able to converge back to the desired reference trajectory in one to two steps while still satisfying state and input constraints. Incorporating optimization over the contact times into the MPC program is left as important future work, which would combine the benefits of both approaches. A depiction of the disturbance rejection behavior of the MPC method can be seen in Fig.~\ref{fig:simdata}.

The MPC with a heuristic reference trajectory and a horizon length of 1.0 second, and the MPC with an HZD trajectory and a horizon length of 0.5 seconds were then deployed on the AMBER hardware. As seen in Fig.~\ref{fig:gait_tiles}, both methods produce forward walking and have a visually distinct gait. We see in the joint angle trajectory data that the MPC with HZD method displays strong periodic behavior, similar to periodic motions expected with an HZD approach.

\section{Conclusion and Outlook} 
\label{sec:conc}

In this work, we proposed a whole-body nonlinear MPC framework that enables online gait optimization using the full rigid body dynamics of a bipedal system. The viability of the presented control structure was shown in simulation and on hardware in a variety of robust dynamic behaviors, including standing, stepping in place, and walking. The addition of a trajectory tracking cost around an offline generated HZD reference enabled similarly robust locomotion at a significantly shorter planning horizon when compared with a heuristic reference or no reference. Motivated by the experimental results and promising reduction in computational complexity, future work will investigate the theoretical properties of using HZD trajectories as terminal components, as well as extensions to 3D walking bipeds.

\bibliographystyle{IEEEtran}
\balance
\bibliography{main}

\end{document}